\documentclass[11pt]{article}
\usepackage[utf8]{inputenc}
\usepackage[T1]{fontenc}
\usepackage{times}
\usepackage{graphicx}
\usepackage{url}
\usepackage{hyperref}
\usepackage{amsmath}
\usepackage{enumitem}
\usepackage{booktabs}
\usepackage{geometry}
\title{Zero-Shot, But at What Cost?\\ Unveiling the Hidden Overhead of MILS’s LLM-CLIP Framework for Image Captioning}
\author{
  \makebox[\textwidth][c]{%
    \parbox{0.45\textwidth}{\centering Yassir Benhammou\\NstarX\\\texttt{yassir.benhammou@nstarxinc.com}}
    \hfill
    \parbox{0.45\textwidth}{\centering Alessandro Tiberio\\NstarX\\\texttt{alessandro.tiberio@nstarxinc.com}}
  }\\[2em]
  \makebox[\textwidth][c]{%
    \parbox{0.45\textwidth}{\centering Gabriel Trautmann\\NstarX\\\texttt{gabriel.trautmann@nstarxinc.com}}
    \hfill
    \parbox{0.45\textwidth}{\centering Suman Kalyan\\NstarX\\\texttt{suman.kalyan@nstarxinc.com}}
  }
}
\date{}

\begin{document}
\maketitle

\begin{abstract}
MILS (Multimodal Iterative LLM Solver) is a recently published framework that claims “LLMs can see and hear without any training” by leveraging an iterative, LLM-CLIP based approach for zero-shot image captioning. While this MILS approach demonstrates good performance, our investigation reveals that this success comes at a hidden, substantial computational cost due to its expensive multi-step refinement process. In contrast, alternative models such as BLIP‑2 and GPT‑4V achieve competitive results through a streamlined, single-pass approach. We hypothesize that the significant overhead inherent in MILS’s iterative process may undermine its practical benefits, thereby challenging the narrative that zero-shot performance can be attained without incurring heavy resource demands. This work is the first to expose and quantify the trade-offs between output quality and computational cost in MILS, providing critical insights for the design of more efficient multimodal models.
\end{abstract}

\section{Introduction}

Recent advances in large-scale pre-training have fundamentally transformed the landscape of multimodal learning. Models such as CLIP \cite{radford2021learning} and the Vision Transformer (ViT) \cite{dosovitskiy2020image} have demonstrated that robust visual representations can be learned from vast amounts of image–text pairs. These breakthroughs, along with pioneering work on large-scale language models \cite{raffel2020exploring, dubey2024llama}, have spurred the development of systems that bridge the gap between vision and language, thereby enabling tasks like image captioning \cite{karpathy2015deep}, video description, and audio captioning in zero-shot or minimally supervised settings.

A particularly intriguing line of research is the exploration of zero-shot methods that do not require task-specific fine-tuning. MILS (Multimodal Iterative LLM Solver) \cite{mils} epitomizes this approach by leveraging the innate reasoning capabilities of large language models (LLMs) together with off-the-shelf multimodal models. MILS employs an iterative, test-time optimization process in which a pre-trained LLM, acting as a \emph{Generator}, produces an extremely diverse set of candidate outputs—on the order of tens of thousands per image—via varied prompts derived from class labels and other textual templates. These candidates are then evaluated by a \emph{Scorer} that uses a CLIP-like model to compute cosine similarity scores between the image's visual features and the candidate captions’ textual representations \cite{radford2021learning}. The top candidates are fed back into the LLM to refine the caption iteratively until convergence or until a preset number of iterations is reached.

Although MILS achieves state-of-the-art performance in zero-shot settings for image captioning, its approach raises critical efficiency concerns. The claim that ``LLMs can see and hear without any training'' is compelling at first glance; however, a deeper examination reveals that MILS relies on a highly resource-intensive iterative process. Generating an initial candidate set of approximately 30K captions per image and refining them over multiple iterations incurs substantial computational overhead, which may negate the practical benefits of its training-free paradigm when compared to more efficient single-pass methods.

In contrast, recent models such as BLIP-2 \cite{blip2} and emerging evaluations of GPT-4V(e.g., \cite{openai_gpt4}) have demonstrated competitive zero-shot performance using a single forward pass. For example, BLIP-2 combines a frozen Vision Transformer (typically ViT-L/14 \cite{dosovitskiy2020image}), a lightweight Query Transformer, and a pre-trained language model (e.g., FLAN-T5-XL \cite{raffel2020exploring}) to directly generate captions, significantly reducing inference time and resource usage. Moreover, chain-of-thought prompting techniques \cite{wei2022chain, kojima2022zero} have further illustrated that large language models can perform complex reasoning without heavy fine-tuning, underscoring the potential for efficient, zero-shot approaches.

This dichotomy prompts a fundamental question: Can the substantial computational overhead inherent in MILS’s iterative process be justified by its performance gains, or do more efficient single-pass models provide a better trade-off between caption quality and resource utilization? In this work, we critically evaluate these trade-offs through extensive experiments on the MSCOCO dataset, measuring caption quality via BLEU, ROUGE\_L, CIDEr, and METEOR, while also profiling computational resources such as inference time and GPU utilization.

The remainder of this paper is structured as follows: Section 2 reviews related work and situates our study in the context of recent advances in multimodal learning and zero-shot techniques; Section 3 details the technical architectures of MILS, BLIP-2, and GPT-4V; Section 4 describes our experimental setup and evaluation metrics; Section 5 presents and discusses our results; and finally, Section 6 concludes with insights and potential directions for future work.

\section{Related Work}

The evolution of image captioning and multimodal learning can be traced through a series of breakthrough models and techniques over the past decade. Early works such as \emph{Show and Tell} \cite{vinyals2015show} and \emph{Show, Attend and Tell} \cite{xu2015show}, both published in 2015, pioneered the integration of convolutional neural networks and recurrent architectures to generate natural language descriptions for images. These foundational methods introduced the notion of learning visual-semantic alignments, paving the way for subsequent advancements. By 2017, researchers introduced more sophisticated training strategies, exemplified by Self-Critical Sequence Training \cite{rennies2017self}, which directly optimized captioning metrics through reinforcement learning, and by 2018, Bottom-Up and Top-Down Attention \cite{anderson2018bottom} further refined caption generation by leveraging object-level features. These innovations greatly improved the fluency and relevance of generated captions. The advent of transformer architectures marked a significant turning point in the field. In 2020, Dosovitskiy et al. \cite{dosovitskiy2020image} introduced the Vision Transformer (ViT), which demonstrated that self-attention can capture high-level semantic information in images without relying on convolutional operations. Shortly thereafter, OpenAI’s CLIP \cite{radford2021learning} leveraged massive-scale image–text pairs to learn joint embeddings, making it possible for models to bridge visual and linguistic modalities in a robust manner. Building on these transformer-based advances, models such as UNITER \cite{chen2020uniter} and Oscar \cite{li2020oscar} (both 2020) further unified visual and textual representations. ALIGN \cite{jing2021align} and ViLT \cite{kim2021vilt}, published in 2021, continued this trend by simplifying architectures while scaling up the training process. These methods enabled significant improvements across a variety of vision and language tasks, from image captioning to visual question answering. Other approaches, such as ZeroCap \cite{tewel2022zerocap} (2022) and MeaCap \cite{zeng2024meacap} (2024), have explored gradient-based optimization or memory-augmented techniques to generate captions without extensive task-specific training. However, these methods often introduce additional complexities or computational costs.

A recent and particularly intriguing development is the shift towards zero-shot methods that eliminate the need for task-specific fine-tuning. MILS (Multimodal Iterative LLM Solver) \cite{mils}, released in 2025, leverages the inherent reasoning capabilities of large language models by employing an iterative, test-time optimization process. In MILS, a pre-trained LLM (the \emph{Generator}) produces an extensive candidate set—approximately 30K candidate captions per image—through diverse prompts. These candidates are then evaluated by a CLIP-like model (the \emph{Scorer}) that computes cosine similarity between the visual features extracted by a frozen encoder (typically ViT-L/14) and the text embeddings. The top candidates are fed back into the LLM for iterative refinement, ultimately converging to a final caption. Although MILS achieves state-of-the-art performance in a zero-shot setting, its iterative process imposes significant computational overhead. In contrast, other recent models such as BLIP-2 \cite{blip2} (2023) and evaluations of GPT-4V\cite{openai_gpt4} have demonstrated that high-quality image captioning can be realized in a single pass. BLIP-2, for instance, integrates a frozen visual encoder with a lightweight query transformer and a pre-trained language model (FLAN-T5-XL), thereby drastically reducing inference time and resource consumption. Similarly, GPT-4’s multimodal capabilities illustrate that it is possible to achieve competitive performance without iterative refinement.

Overall, the literature reveals not only a trajectory of continual improvements in caption quality and multimodal understanding but also a growing concern regarding the computational resources required by increasingly complex models. Our work is motivated by this dual challenge: to assess whether the substantial computational overhead of iterative zero-shot methods like MILS is justified by their performance gains compared to more efficient, single-pass alternatives such as BLIP-2 and GPT-4.

\section{Evaluated Models and their Technical Methodologies}
In this section, we present the technical methodologies underlying the evaluated image captioning models. The following subsections provide detailed descriptions of the workflows implemented by each model. First, we outline the MILS workflow, which employs an iterative process to refine candidate captions generated by a large language model using a CLIP-like scoring mechanism. Next, we describe the BLIP-2 workflow, which integrates a frozen visual encoder, a lightweight query transformer, and a pre-trained language model to produce captions in a single forward pass. Finally, we detail the GPT-4V workflow, where captions are generated through a direct API call in a streamlined, one-pass manner. Together, these descriptions illustrate the diverse strategies adopted for image captioning and set the stage for our subsequent evaluation.

\subsection{MILS}
MILS (Multimodal Iterative LLM Solver)\cite{mils} is a training‑free framework that augments the zero‑shot capabilities of large language models (LLMs) for a variety of multimodal tasks such as image, video, and audio captioning as well as text‑to‑image generation and style transfer \cite{mils}. At its core, MILS comprises two primary components: the Generator and the Scorer. The Generator utilizes a pre‑trained LLM (e.g., Llama 3.1 8B \cite{dubey2024llama}) to produce an exceptionally diverse initial set of candidate text outputs by leveraging varied prompts derived from class labels and other textual templates—resulting in roughly tens of thousands (approximately 30K per image) of candidate captions. This enormous candidate pool is intended to maximize diversity and ensure sufficient coverage of potential descriptions. In contrast, the Scorer employs a CLIP‑like model \cite{radford2021learning} featuring a frozen visual encoder (typically a ViT‑L/14 \cite{dosovitskiy2020image}) and a corresponding text encoder. The Scorer encodes both the input image and each candidate caption into a shared embedding space and computes cosine similarity scores to determine how well each caption aligns with the image content. MILS then operates in an iterative optimization loop: the highest‑scoring candidates (for instance, the top 50) are fed back into the Generator as feedback to prompt refined text generations. This loop continues until convergence or until a preset iteration limit is reached, ultimately yielding a final caption that best captures the essence of the input image. While this modular design enables MILS to generalize across multiple modalities and tasks, our analysis reveals that the iterative candidate generation and refinement process imposes a substantial computational overhead, thereby challenging the claim that zero‑shot performance can be achieved “without any training” \cite{mils, radford2021learning}.

\subsection{BLIP-2}
BLIP‑2 is built upon three core components that jointly enable efficient zero-shot image captioning. The first component is a frozen Vision Transformer (ViT‑L/14), pre-trained on large-scale image datasets such as ImageNet, which extracts rich visual features from input images \cite{dosovitskiy2020image}. These features are then processed by a lightweight Query Transformer (Q‑Former) that acts as an intermediary, transforming the high-dimensional visual representations into tokens that the language model can interpret. Finally, the system leverages a large, frozen language model—the FLAN‑T5‑XL variant—which has been extensively pre-trained on vast text corpora and fine‑tuned via instruction tuning to generate coherent and contextually relevant captions \cite{raffel2020exploring, flan}. Together, these components allow BLIP‑2 to perform image captioning in a single forward pass, achieving competitive zero-shot performance without any additional task-specific training \cite{blip2}.

\subsection{GPT-4V}
Our GPT-4V workflow leverages OpenAI's GPT-4VV multimodal API to generate image captions in a single, non-iterative pass. In this approach, the process begins by loading test images (from the MSCOCO validation set) along with their corresponding split information from a JSON file. Each image is first opened, converted to RGB, and then encoded into a base64 string. This encoded image is embedded as a data URL in a JSON payload along with a predefined text prompt (e.g., "Describe this image concisely in one sentence."). The payload is then sent via an HTTP POST request to the GPT-4VV API endpoint with the appropriate API key for authentication.

Upon receiving a response, the generated caption is extracted from the API's output and saved to a log file for each image, while also being aggregated into a JSON file to facilitate further evaluation. Rate limiting is applied between API requests to manage throughput and avoid exceeding rate limits. By directly generating the caption in a single API call, GPT-4VV bypasses the iterative refinement stage seen in MILS, thereby dramatically reducing computational overhead while leveraging its advanced multimodal understanding to produce high-quality captions \cite{openai_gpt4}. This streamlined, one-pass approach is contrasted with the iterative, feedback-driven processes of MILS and serves as an efficient alternative for zero-shot image captioning.

\section{Experimental Setup}
In this section, we detail the experimental setup used to evaluate our zero-shot image captioning models. We first describe the dataset utilized in this study, followed by an overview of the evaluation metrics employed to assess both caption quality and computational efficiency. Finally, we outline the implementation details of each model, including MILS, BLIP-2, and GPT-4V, highlighting the hardware and runtime environments used for their execution. The subsequent subsections provide the necessary references and further contextualize our experimental approach.
\subsection{Dataset}
Microsoft COCO (Common Objects in Context) is a widely adopted benchmark for computer vision tasks, including object detection, segmentation, and image captioning. The 2014 version of MSCOCO contains over 82,000 images for training and more than 40,000 images for validation, with each image annotated by multiple human-generated captions (typically five per image) \cite{lin2014microsoft}. For our experiments, we use the MSCOCO 2014 validation set, which consists of 5,000 images. This set is extensively used in the literature for evaluating image captioning models \cite{karpathy2015deep}, as it provides a diverse range of visual scenes and rich contextual descriptions, making it an ideal resource to benchmark both caption quality and the efficiency of zero-shot models. The dataset’s high-quality annotations enable rigorous evaluation using standard metrics such as BLEU and METEOR, facilitating a meaningful comparison between different models evaluated in this study.

\subsection{Metrics}
We evaluate the generated captions using a comprehensive suite of metrics based on the COCO evaluation framework (implemented in \texttt{pycocoevalcap}) and the Hugging Face \texttt{evaluate} library. Specifically, our quantitative assessment includes:

\begin{itemize}[noitemsep]
    \item \textbf{BLEU:} We report BLEU-1, BLEU-2, BLEU-3 and BLEU-4 scores \cite{papineni2002bleu} to measure the n-gram precision between the generated captions and human-annotated references.
    \item \textbf{ROUGE\_L:} This metric evaluates the longest common subsequence overlap, capturing recall-based quality in the text.
    \item \textbf{CIDEr:} The CIDEr metric \cite{vedantam2015cider} computes TF-IDF weighted consensus between the generated and reference captions, emphasizing the relevance of content.
    \item \textbf{METEOR:} We also use the METEOR score \cite{banerjee2005meteor} (via the Hugging Face \texttt{evaluate} library) to assess semantic similarity, considering synonyms and stemmed variants.
\end{itemize}

In addition to these caption quality metrics, we record computational efficiency metrics such as inference time and GPU utilization. This dual evaluation allows us to weigh the trade-offs between output performance and resource consumption across the different models.

\subsection{Implementation}
MILS is implemented as described in \cite{mils}. BLIP-2 and GPT-4V are run in zero-shot mode using their public implementations. All experiments are performed on comparable hardware.
\begin{itemize}[noitemsep]
    \item \textbf{MILS:} The original research paper spread the workload across 8 Nvidia A100 GPUs, each operating separately from the other \cite{mils}. For our replication, we split the data into 50 chunks of 100 images. As the chunk size had no bearing on the final outcome, this method was beneficial in avoiding unexpected errors and minimizing additional expenditure. We ran these chunks on the same A100-80GB GPUs, which we accessed through Modal. No changes were made to the original MILS code provided in the git repository; only a wrapper was added to the MILS image captioning script so that it could be executed on Modal. 

    It is worth noting that we occasionally encountered out-of-memory errors. All metrics and hardware configurations were identical to the original MILS experiment, so these errors were unexpected. To validate our replication, we ensured that each chunk produced 100 \texttt{log.txt} files, each containing 11 responses (one original response sourced from the candidate list of approximately \mbox{30K} captions and 10 iterative responses).

    \item \textbf{BLIP-2:} BLIP-2 is run in zero-shot inference mode using its public implementation \cite{blip2} on Modal with one Nvidia A10 GPU. To deploy on Modal, we created a wrapper that builds the corresponding image containing the entire script. In this setup, each image is processed in a single forward pass by a model that integrates a frozen Vision Transformer with a lightweight Query Transformer and a pre-trained language model. This efficient architecture obviates the need for iterative refinement. Generated captions for this model were saved in a JSON file, \texttt{Blip2\_5000.json}.
    
    \item \textbf{GPT-4V:} For GPT-4V, we attempted to closely mimic the initial MILS execution process while removing the iterative refinement. We used essentially the same first line from the MILS prompt—omitting the additional instructions related to iteration. GPT-4V processed one image at a time and produced a single-sentence caption. The resulting captions were saved in the same directory structure as MILS, with each image’s caption stored in a \texttt{log.txt} file within a directory named after its respective COCO image ID.
\end{itemize}
For reproducibility and further exploration, the complete implementation—including the original models code and our added wrappers for execution on Modal—is publicly available at our \href{https://github.com/ChambaPie/MILS_Benchmarking.git}{GitHub Repo}. Additionally, the used images and partitions with the generated caption outputs for each evaluated model (MILS, BLIP-2, and GPT-4V) are stored in our shared \href{https://drive.google.com/drive/folders/1jR33P2jbj3e6Uy9wuk2d0G3ljPFl6ZqJ?usp=sharing}{Drive Folder}.

\section{Results and Discussion}

In this section, we present a comprehensive evaluation of the three image captioning models. Our quantitative analysis measures caption quality using standard metrics such as BLEU, METEOR, ROUGE\_L, and CIDEr on the MSCOCO dataset. In addition, we report computational efficiency metrics including running time, GPU usage, and cost to capture the resource demands of each model. Complementing the numerical results, our qualitative evaluation visually contrasts the captions generated by GPT-4V, BLIP-2, and MILS, highlighting differences in descriptive detail, linguistic fluency, and contextual grounding. This integrated evaluation provides new insights into the trade-offs between output quality and computational efficiency in the evaluated approaches.

Table~\ref{tab:caption_metrics} summarizes the caption quality metrics achieved on the MSCOCO 2014 validation set for the three evaluated models: GPT-4V, BLIP-2, and MILS. For instance, BLIP-2 attains the highest BLEU-1 (0.6713) and CIDEr (0.8569) scores, while GPT-4VV follows with BLEU-1 at 0.5839 and CIDEr at 0.5917. In contrast, MILS produces considerably lower values across all metrics (e.g., BLEU-1: 0.4538, CIDEr: 0.3231). BLEU-2, BLEU-3, BLEU-4, and METEOR scores exhibit a similar trend, and although ROUGE\_L is slightly higher for GPT-4V, the overall caption quality from BLIP-2 is notably superior.

\begin{table}[ht]
\centering
\caption{Caption Quality Metrics on MSCOCO 2014 (Validation Set)}
\label{tab:caption_metrics}
\begin{tabular}{lccc}
\toprule
\textbf{Metric}  & \textbf{GPT-4V} & \textbf{BLIP-2} & \textbf{MILS} \\ \midrule
BLEU-1           & 0.5839         & 0.6713         & 0.4538      \\
BLEU-2           & 0.4010         & 0.5161         & 0.2666      \\
BLEU-3           & 0.2677         & 0.3797         & 0.1437      \\
BLEU-4           & 0.1748         & 0.2723         & 0.0778      \\
METEOR           & 0.2726         & 0.4662         & 0.1496      \\
ROUGE\_L         & 0.4744         & 0.4700         & 0.3130      \\
CIDEr            & 0.5917         & 0.8569         & 0.3231      \\
\bottomrule
\end{tabular}
\end{table}

Table~\ref{tab:execution_cost} presents a comparison of the execution cost for the three models. GPT-4VV processes the dataset in approximately 6 hours and 48 minutes at a cost of \textdollar{}5.41, and BLIP-2 is even more efficient, running in roughly 31 minutes and 34 seconds at a modal cost of \textdollar{}0.65. In stark contrast, MILS requires a staggering runtime of 116 hours and 40 minutes, incurring a cost of \textdollar{}291.66. These figures underscore the heavy computational overhead of MILS's iterative approach.

\begin{table}[ht]
\centering
\caption{Execution Cost and Hardware Environment Comparison}
\label{tab:execution_cost}
\begin{tabular}{lccc}
\toprule
\textbf{Metric/Environment} & \textbf{GPT-4V} & \textbf{BLIP-2} & \textbf{MILS} \\ \midrule
Running Time   & 6h 48m         & 31m 34s       & 116h 40m     \\
Cost           & \textdollar{}5.41 & \textdollar{}0.65 & \textdollar{}291.66 \\ 
GPU Used       & OpenAI server     & Nvidia A10     & 8 Nvidia A100-80GB \\
Runtime Env.   & Locally(Mac OS) & Modal (Debian Slim) & Modal (Debian Slim) \\ \bottomrule
\end{tabular}
\end{table}

The contrasting results reveal a pronounced trade-off between caption quality and computational efficiency. While BLIP-2 and GPT-4V, both employing a single-pass inference strategy, generate high-quality captions with modest resource consumption, MILS’s iterative refinement—despite its potential to further boost quality—incurs significant computational overhead. This work thereby challenges the assertion that zero-shot performance can be achieved “without any training,” as MILS’s seemingly training-free iterative optimization is coupled with a heavy cost in terms of runtime and resource usage. In summary, our findings indicate that, although MILS demonstrates competitive performance in certain quality metrics, its extensive iterative process makes it less practical compared to the more efficient, single-pass alternatives like BLIP-2 and GPT-4V, especially in resource-constrained environments. This suggests that the hidden overhead of MILS’s iterative process may not justify its performance gains, challenging its claim of ``no training.''

\begin{figure}[ht]
\centering
\includegraphics[width=\textwidth]{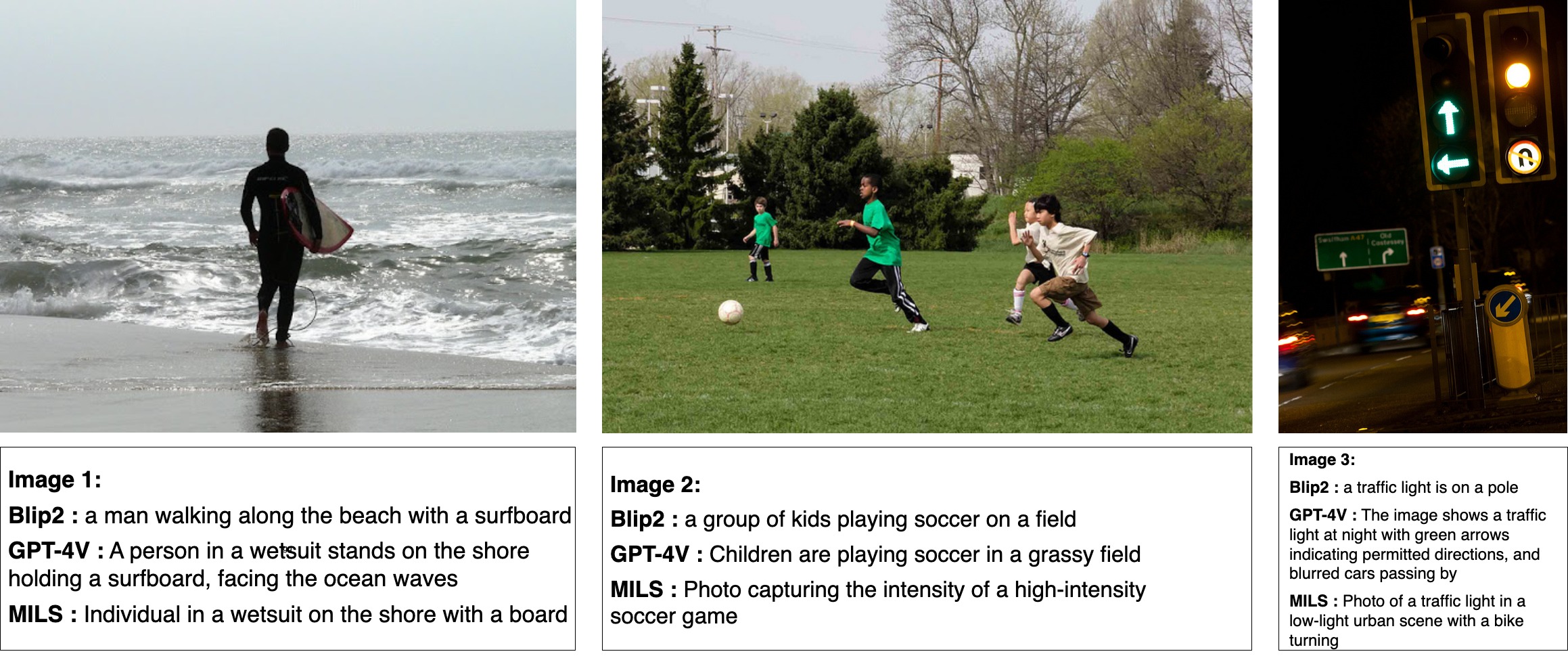}
\caption{Qualitative Comparison of Captions. Each column corresponds to one image from the COCO dataset, and shows the caption outputs generated by GPT-4V, BLIP-2, and MILS. The figure highlights differences in descriptive detail, linguistic fluency, and contextual grounding among the three approaches.}
\label{fig:qualitative}
\end{figure}

To complement the quantitative evaluation, we present in figure ~\ref{fig:qualitative} a qualitative comparison of the captions generated by MILS, BLIP-2, and GPT-4V for three diverse samples from the COCO dataset. Each column displays one image alongside the corresponding caption outputs from each model, highlighting the stylistic and descriptive differences in how each model interprets visual content. For the first image featuring a surfer on the beach, GPT-4V produced the most detailed and natural description, identifying both attire and contextual background, whereas BLIP-2 offered a concise and accurate summary; MILS recognized the essential elements but lacked linguistic fluency and expressiveness. In the second image, which depicts children playing soccer, both GPT-4V and BLIP-2 provided fluent and grounded captions, while MILS generated a redundant and more abstract description. The third image, showing a nighttime traffic scene, further underscores these differences: GPT-4V excelled in capturing contextual and dynamic details, BLIP-2 produced a minimal caption that omitted critical environmental cues, and MILS delivered a more interpretive description that occasionally included hallucinated elements (e.g., an extraneous mention of a bike). Overall, GPT-4V consistently demonstrated superior scene understanding and descriptive clarity, BLIP-2 balanced accuracy with brevity, and MILS, though interpretable, lagged in human-likeness and contextual depth due to its computationally intensive iterative architecture.

\section{Conclusion}
In this paper, we conducted a comprehensive evaluation of three zero-shot image captioning models—MILS, BLIP-2, and GPT-4V on the MSCOCO 2014 validation set. Our quantitative analysis using BLEU, METEOR, ROUGE\_L, and CIDEr metrics, coupled with a detailed examination of execution costs, revealed a pronounced trade-off between caption quality and computational efficiency.

While MILS’s iterative approach is innovative, enabling high-quality caption refinement through repeated feedback loops, our experiments show that this method incurs significant computational overhead. In contrast, BLIP-2 and GPT-4V, which generate captions in a single forward pass, not only provide competitive caption quality but do so at a fraction of the runtime and cost. The qualitative comparisons further illustrate that GPT-4V consistently offers superior descriptive clarity and scene understanding, with BLIP-2 balancing brevity and accuracy, whereas MILS struggles with contextual depth and occasional hallucinations.

These findings challenge the assertion that zero-shot performance can be achieved “without any training,” as MILS’s so-called training-free approach effectively substitutes explicit training with an expensive iterative optimization process. Our study highlights the importance of considering both performance and resource efficiency when designing multimodal systems—an essential factor in real-world applications, particularly in resource-constrained environments.

Future work will explore hybrid approaches that combine the strengths of iterative refinement with the efficiency of single-pass inference, as well as optimization techniques to reduce the overhead associated with iterative methods. Overall, our work provides valuable insights and establishes a foundation for developing more practical and scalable zero-shot image captioning systems.

\end{document}